\documentclass[10pt,twocolumn,letterpaper]{article}

\usepackage{cvpr}
\usepackage{times}
\usepackage{epsfig}
\usepackage{graphicx}
\usepackage{amsmath}
\usepackage{amssymb}
\usepackage[pagebackref=true,breaklinks=true,letterpaper=Kangle’s paper,colorlinks,bookmarks=false]{hyperref}

\usepackage[breaklinks=true,bookmarks=false]{hyperref}

\cvprfinalcopy 

\def\cvprPaperID{****} 

\begin{document}

\title{Multi-Oriented Text Detection with Fully Convolutional Networks}
\author{Zheng Zhang$^{1*}$~~~Chengquan Zhang$^1$\thanks{\scriptsize Authors contributed equally}~~~Wei Shen$^2$~~~Cong Yao$^1$~~~Wenyu Liu$^1$~~Xiang Bai$^1$\\
$^1$ School of Electronic Information and Communications, Huazhong University of Science and Technology\\
$^2$ Key Laboratory of Specialty Fiber Optics and Optical Access Networks, Shanghai University\\
{\tt\small macaroniz1990@gmail.com, zchengquan@gmail.com, wei.shen@t.shu.edu.cn} \\
{\tt\small yaocong2010@gmail.com, liuwy@mail.hust.edu.cn, xbai@hust.edu.cn}
}

\maketitle

\begin{abstract}
In this paper, we propose a novel approach for text detection in natural images. Both local and global cues are taken into account for localizing text lines in a coarse-to-fine procedure. First, a Fully Convolutional Network (FCN) model is trained to predict the salient map of text regions in a holistic manner. Then, text line hypotheses are estimated by combining the salient map and character components. Finally, another FCN classifier is used to predict the centroid of each character, in order to remove the false hypotheses. The framework is general for handling text in multiple orientations, languages and fonts. The proposed method consistently achieves the state-of-the-art performance on three text detection benchmarks: MSRA-TD500, ICDAR2015 and ICDAR2013.
\end{abstract}

\section{Introduction}
Driven by the increasing demands for many computer vision tasks, reading text in the wild (from scene images) has become an active direction in this community. Though extensively studied in recent years, text spotting under uncontrolled environments is still quite challenging. Especially, detecting text lines with arbitrary orientations is an extremely difficult task, as it takes much more hypotheses into account, which  drastically enlarges the searching space. Most existing approaches are successfully designed for detecting horizontal or near-horizontal text~\cite{Ref:Chen2004,Ref:Epshtein2010,Ref:Neumann2010,Ref:Neumann2013B,Ref:Bissacco2013,Ref:Jaderberg2014,Ref:Huang2014,zhang2015symmetry,tiantext,qin2016fast,xiongtext}. However, there is still a large gap when applying them to multi-oriented text, which has been verified by the low accuracies reported in the recent ICDAR2015 competition for text detection ~\cite{karatzasicdar}.  

\begin{figure}[!h]
\begin{center}
\includegraphics[width=0.9\linewidth]{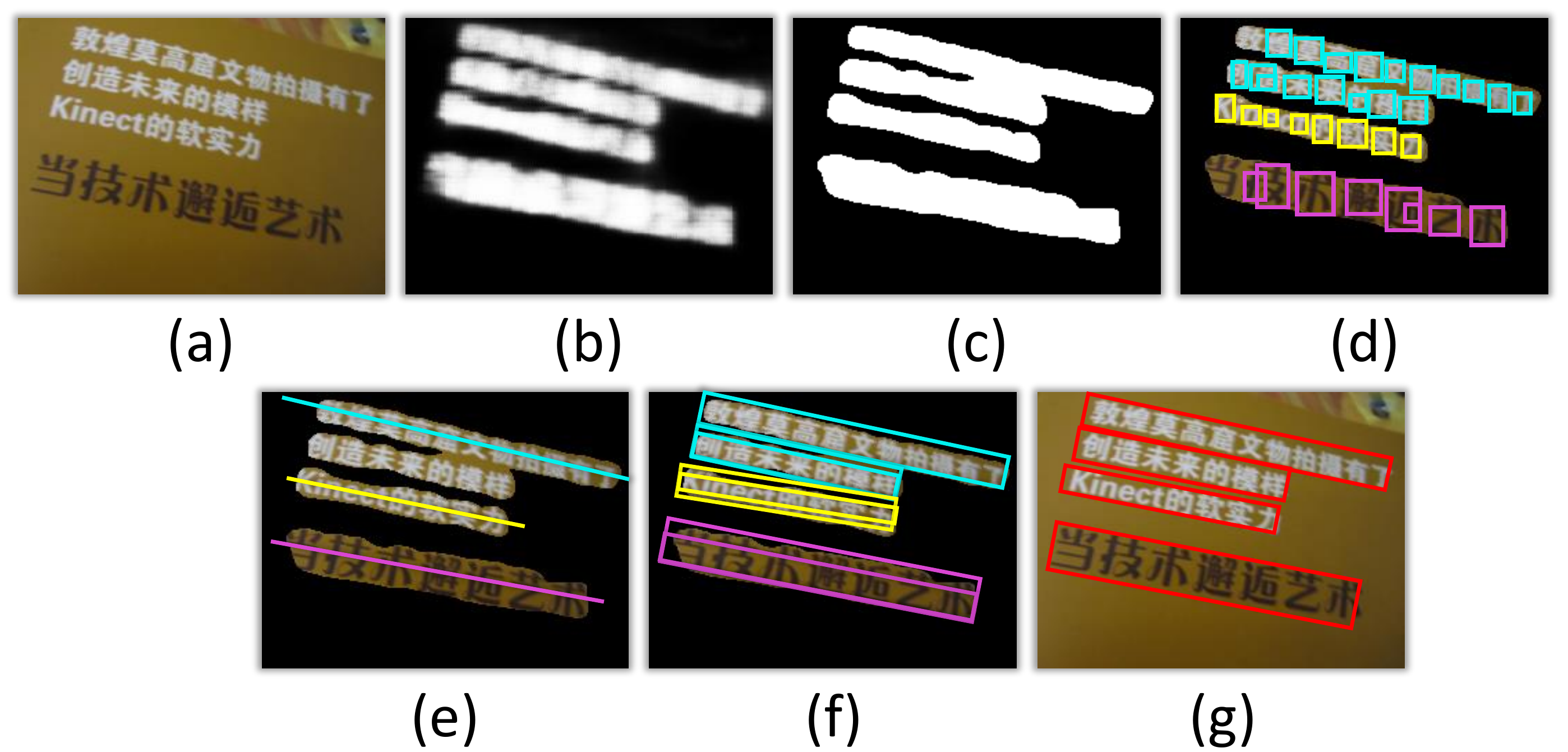}
\end{center}
\vspace{-5mm}
\caption{The procedure of the proposed method. (a) An input image; (b) The salient map of the text regions predicted by the Text-Block FCN; (c) Text block generation; (d) Candidate character component extraction; (e) Orientation estimation by component projection; (f) Text line candidates extraction; (g) The detection results of the proposed method.}
\label{fig:introduction_pipeline}
\vspace{-2mm}
\end{figure}

Text, which can be treated as sequence-like objects with unconstrained lengths, possesses very distinctive appearance and shape compared to generic objects. Consequently, the detection methods in scene images based on sliding windows ~\cite{Ref:Jaderberg2014,Ref:Wang2012ICPR,Ref:Chen2004,Ref:Wang2010,Ref:Neumann2013B} and connected component~\cite{Ref:NeumannM12MSER,Ref:Huang2014,Ref:Epshtein2010,yinmulti,Ref:Yao2014C} have become mainstream in this specific domain. In particular, the component-based methods utilizing Maximally Stable Extremal Regions (MSER) ~\cite{Ref:Neumann2010} as the basic representations achieved the state-of-the-art performance on ICDAR2013 and ICDAR2015 competitions ~\cite{Ref:Karatzas2013,karatzasicdar}. Recently, ~\cite{Ref:Huang2014} utilized a convolution neural network to learn highly robust representations of character components. Usually, the component grouping algorithms including clustering algorithms or some heuristic rules are essential for localizing text at a word or line level. As an unconventional approach,~\cite{zhang2015symmetry} directly hits text lines from cluttered images, benefiting from symmetry and self-similarity properties of them. Therefore, it seems that both local (character components) and global (text regions) information are very helpful for text detection.

\begin{figure*}[ht]
\begin{center}
\includegraphics[width=0.9\linewidth]{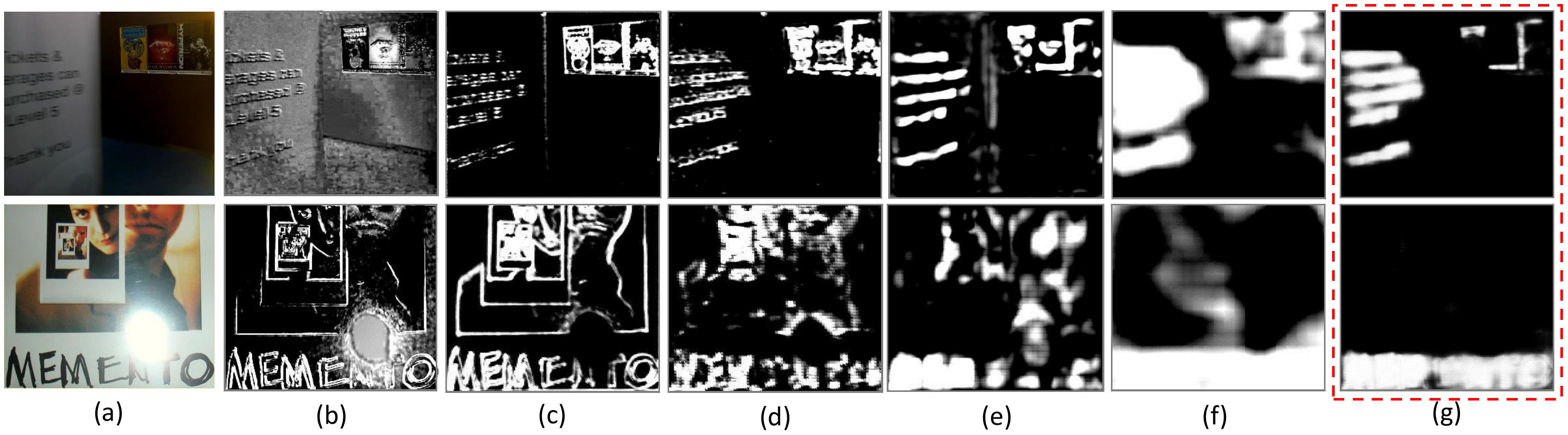}
\end{center}
\caption{The illustration of feature maps generated by the Text-Block FCN. (a) Input images; (b)$\sim$(f) The feature maps from stage$1$$\sim$stage$5$. Lower level stages capture more local structures, and higher level stages capture more global information; (g) The final salient maps.}
\label{fig:feature_maps}
\vspace{-2mm}
\end{figure*}

\begin{figure*}[ht]
\begin{center}
\includegraphics[width=0.9\linewidth]{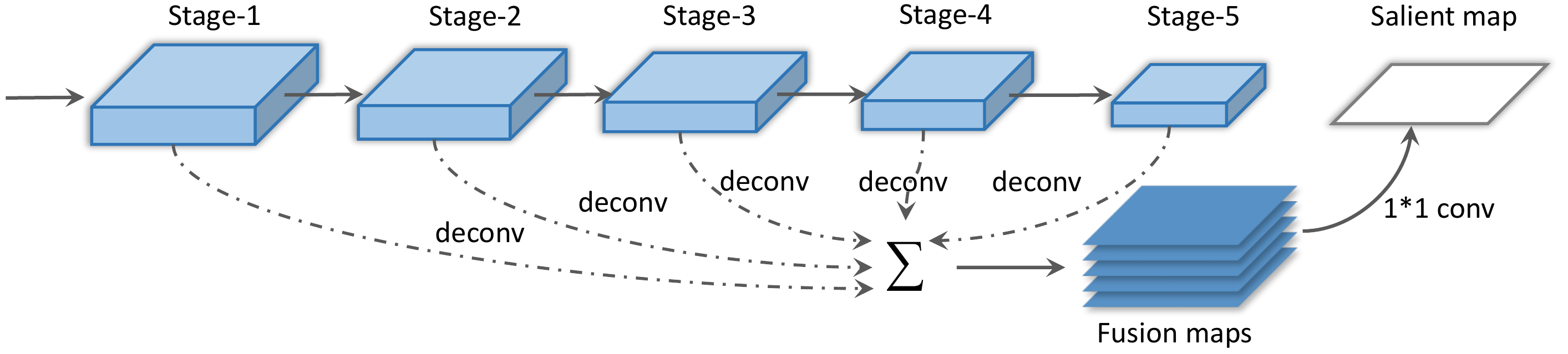}
\end{center}
\caption{The network architecture of the Text-Block FCN whose 5 convolutional stages are inherited from VGG 16-layer model. For each stage, a deconvolutional layer (equals to a $1\times 1$ convolutional layer and a upsampling layer) is connected. All the feature maps are concatenated with a $1\times 1$ convolutional layer and a sigmoid layer.}
\label{fig:fcn}
\vspace{-2mm}
\end{figure*}

In this paper, an unconventional detection framework for multi-oriented text is proposed. The basic idea is to integrate local and global cues of text blocks with a coarse-to-fine strategy. At the coarse level, a pixel-wise text/non-text salient map is efficiently generated by utilizing a Fully Convolutional Network (FCN) ~\cite{long2015fully}. We show that the salient map provides a powerful guidance for estimating orientations and generating candidate bounding boxes of text lines, while combining it with local character components. More specifically, the pipeline of the proposed detection framework is shown in Fig.~\ref{fig:introduction_pipeline}. First, a salient map is generated and segmented into several candidate text blocks. Second, character components are extracted from the text blocks. Third, projections of the character components are used for estimating the orientation. Then, based on the estimated orientation, all candidate bounding boxes of text lines are constructed by integrating cues from the components and the text blocks. Finally, detection results are obtained by removing false candidates through a filtering algorithm.         

Our contributions are in three folds:
First, we present a novel way for computing text salient map, through learning a strong text labeling model with FCN. The text labeling model is trained and tested in a holistic manner, highly stable to the large scale and orientation variations of scene text, and quite efficient for localizing text blocks at the coarse level. In addition, it is also applicable to multi-script text. 
Second, an efficient method for extracting bounding boxes of text line candidates in multiple orientations is presented. We show that the local (character components) and the global (text blocks from the salient map) cues are both helpful and complementary to each other. 
Our third contribution is to propose a novel method for filtering false candidates. We train an efficient model (another FCN) to predict character centroids within the text line candidates. We show that the predicted character centroids provide accurate positions of each character, which are effective features for removing the false candidates. 
The proposed detection framework achieves the state-of-the-art performance on both horizontal and multi-oriented scene text detection benchmarks. 

The remainder of this paper is organized as follows: In Sec.~\ref{Sec:relatedWork}, we briefly review the previously related work. In Sec.~\ref{Sec:methodology}, we describe the proposed method in detail, including text block detection, strategies for multi-oriented text line candidate generation, and false alarm removal. Experimental results are presented in Sec.~\ref{Sec:Experiments}. Finally, conclusion remarks and future work are given in Sec.~\ref{Sec:Conclusion}.


\section{Related Work} \label{Sec:relatedWork}
Text detection in natural images has received much attention from the communities of computer vision and document analysis. However, most text detection methods focus on detecting horizontal or near-horizontal text mainly in two ways: 1) localizing the bounding boxes of words~\cite{Ref:Epshtein2010, Ref:Chen2004, Ref:Neumann2013B, Ref:Neumann2010,Ref:Pan2011, Ref:Yin2014, Ref:Huang2013, Ref:Huang2014}, 2) combining detection and recognition procedures into an end-to-end text recognition method~\cite{Ref:Jaderberg2014,Ref:Yao2014C}. Comprehensive surveys for scene text detection and recognition can be referred to ~\cite{Ref:Ye2014,zhu2016scene}. 

In this section, we focus on the most relevant works that are presented for multi-oriented text detection.
Multi-oriented text detection in the wild is first studied by~\cite{Ref:Yi2011, Ref:Yao2012}. Their detection pipelines are similar to the traditional methods based on connected component extraction, integrating orientation estimation of each character and text line. ~\cite{Ref:Kang2014} treated each MSER component as a vertex in a graph, then text detection is transferred into a graph partitioning problem. ~\cite{yinmulti} proposed a multi-stage clustering algorithm for grouping MSER components to detect multi-oriented text. ~\cite{Ref:Yao2014C} proposed an end-to-end system based on SWT~\cite{Ref:Epshtein2010} for multi-oriented text. Recently, a challenging benchmark for multi-oriented text detection has been released for the ICDAR2015 text detection competition, and many researchers have reported their results on it.  

In addition, it is worth mentioning that both of the recent approaches~\cite{Ref:Wang2012ICPR, Ref:Jaderberg2014, Ref:Huang2014} and our method, which used the deep convolutional neural network, have achieved superior performance over conventional approaches in several aspects: 1) learn a more robust component representation by pixel labeling with CNN~\cite{Ref:Jaderberg2014}; 2) leverage the powerful discrimination ability of CNN for better eliminating false positives~\cite{Ref:Huang2014,zhang2015symmetry}; 3) learn a strong character/word recognizer with CNN for end-to-end text detection~\cite{Ref:Wang2012ICPR,jaderberg2014reading}. However, these methods only focus on horizontal text detection.

\section{Proposed Methodology}\label{Sec:methodology}
In this section, we describe the proposed method in detail. First, text blocks are detected via a fully convolutional network (named Text-Block FCN). Then, multi-oriented text line candidates are extracted from these text blocks by taking the local information (MSER components) into account. Finally, false text line candidates are eliminated by the character centroid information. The character centroid information is provided by a smaller fully convolutional network (named Character-Centroid FCN).

\subsection{Text Block Detection}\label{Sec:textBlock}
\begin{figure}[!h]
\begin{center}
\includegraphics[width=0.9\linewidth]{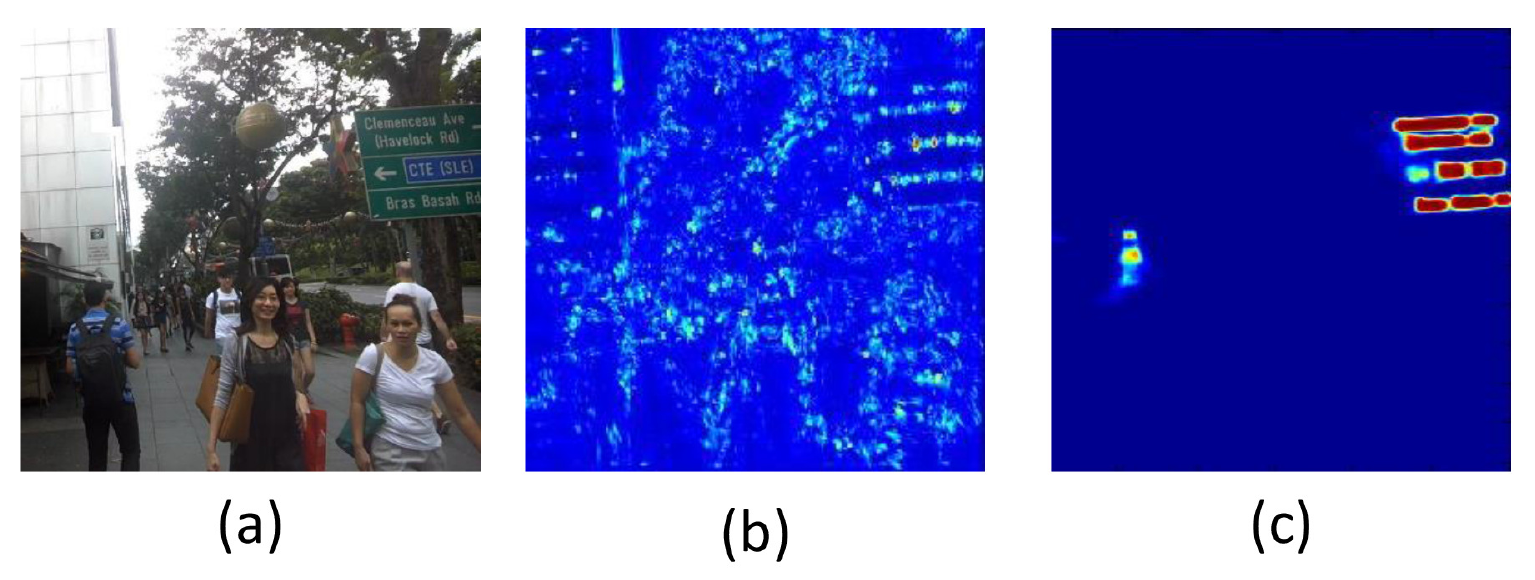}
\end{center}
\vspace{-2mm}
\caption{The results of a character detector and our method. (a) An input image; (b) The character response map, which is generated by the state-of-the-art method~\cite{Ref:Jaderberg2014}; (c) The salient map of text regions, which is generated by the Text-Block FCN.}
\label{fig:comp_map}
\end{figure}

In the past few years, most of the leading methods in scene text detection are based on detecting characters. In early practice~\cite{Ref:NeumannM12MSER,Ref:Pan2011,Ref:Yao2012},  a large number of manually designed features are used to identify characters with strong classifiers. Recently, some works~\cite{Ref:Huang2014,Ref:Jaderberg2014} have achieved great performance, adopting CNN as a character detector. However, even the state-of-the-art character detector~\cite{Ref:Jaderberg2014} still performs poorly at complicated background (Fig.~\ref{fig:comp_map} (b)). The performance of the character detector is limited due to three aspects: firstly, characters are susceptible to several conditions, such as blur, non-uniform illumination, low resolution, disconnected stroke, etc.; secondly, a great quantity of elements in the background are similar in appearance to characters, making them extremely hard to distinguish; thirdly, the variation of the character itself, such as fonts, colors, languages, etc., increases the learning difficulty for classifiers. By comparison, text blocks possess more distinguishable and stable properties. Both local and global appearances of text block are useful cues for distinguishing between text and non-text regions (Fig.~\ref{fig:comp_map} (c)).

Fully convolutional network (FCN), a deep convolutional neural network proposed recently, has achieved great performance on pixel level recognition tasks, such as object segmentation~\cite{long2015fully} and edge detection~\cite{xie2015holistically}. This kind of network is very suitable for detecting text blocks, owing to several advantages: 1) It considers both local and global context information at the same time.; 2) It is trained in an end-to-end manner; 3) Benefiting from the removal of fully connected layers, FCN is efficient in pixel labeling. In this section, we learn a FCN model, named Text-Block FCN, to label salient regions of text blocks in a holistic way.

\textbf{Text-Block FCN} 
We convert the VGG 16-layer net~\cite{simonyan2014very} into our text block detection model that is illustrated in Fig.~\ref{fig:fcn}. The first 5 convolutional stages are derived from the VGG 16-layer net. The receptive field sizes of the convolutional stages are variable, contributing to that different stages can capture context information with different sizes. Each convolutional stage is followed by a deconvolutinal layer (equals to a $1\times 1$ convolutional layer and a upsampling layer) to generate feature maps of the same size. The discriminative and hierarchical fusion maps are then the concatenation in depth of these upsampled maps. Finally, the fully-connected layers are replaced with a $1\times 1$ convolutional layer and a sigmoid layer to efficiently make the pixel-level prediction. 

\begin{figure}[!h]
\begin{center}
\includegraphics[width=0.9\linewidth]{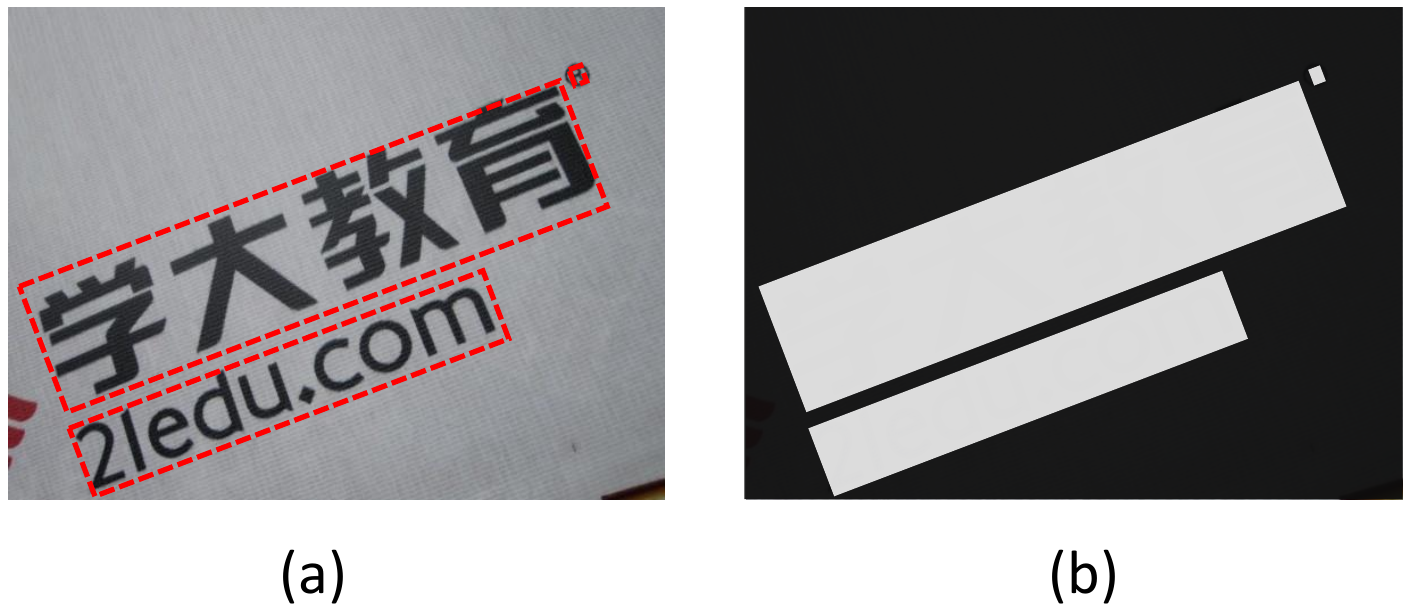}
\end{center}
\vspace{-2mm}
\caption{The illustration of the ground truth map used in the training phase of the Text-Block FCN. (a) An input image. The text lines within the image are labeled with red bounding boxes; (b) The ground truth map.}
\label{fig:groundtruth}
\end{figure}

In the training phase, pixels within the bounding box of each text line or word are considered as the positive region for the following reasons: firstly, the regions between adjacent characters are distinct in contrast to other non-text regions; secondly, the global structure of text can be incorporated into the model; thirdly, bounding boxes of text lines or words are easy to be annotated and obtained. An example of the ground truth map is shown in Fig.~\ref{fig:groundtruth}. The cross-entropy loss function and stochastic gradient descent are used to train this model. 

In the testing phase, the salient map of text regions, leveraging all context information from different stages, is computed by the trained Text-Block FCN model at first. As shown in Fig.~\ref{fig:feature_maps}, the feature map of stage-1 captures more local structures like gradient (Fig.~\ref{fig:feature_maps} (b)), while the higher level stages capture more global information (Fig.~\ref{fig:feature_maps} (e) (f)). Then, the pixels whose probability is larger than $0.2$ are reserved, and the connected pixels are grouped together into several text blocks. An example of the text block detection result is shown in Fig.~\ref{fig:introduction_pipeline} (c).

\begin{figure}[!h]
\begin{center}
\includegraphics[width=0.95\linewidth]{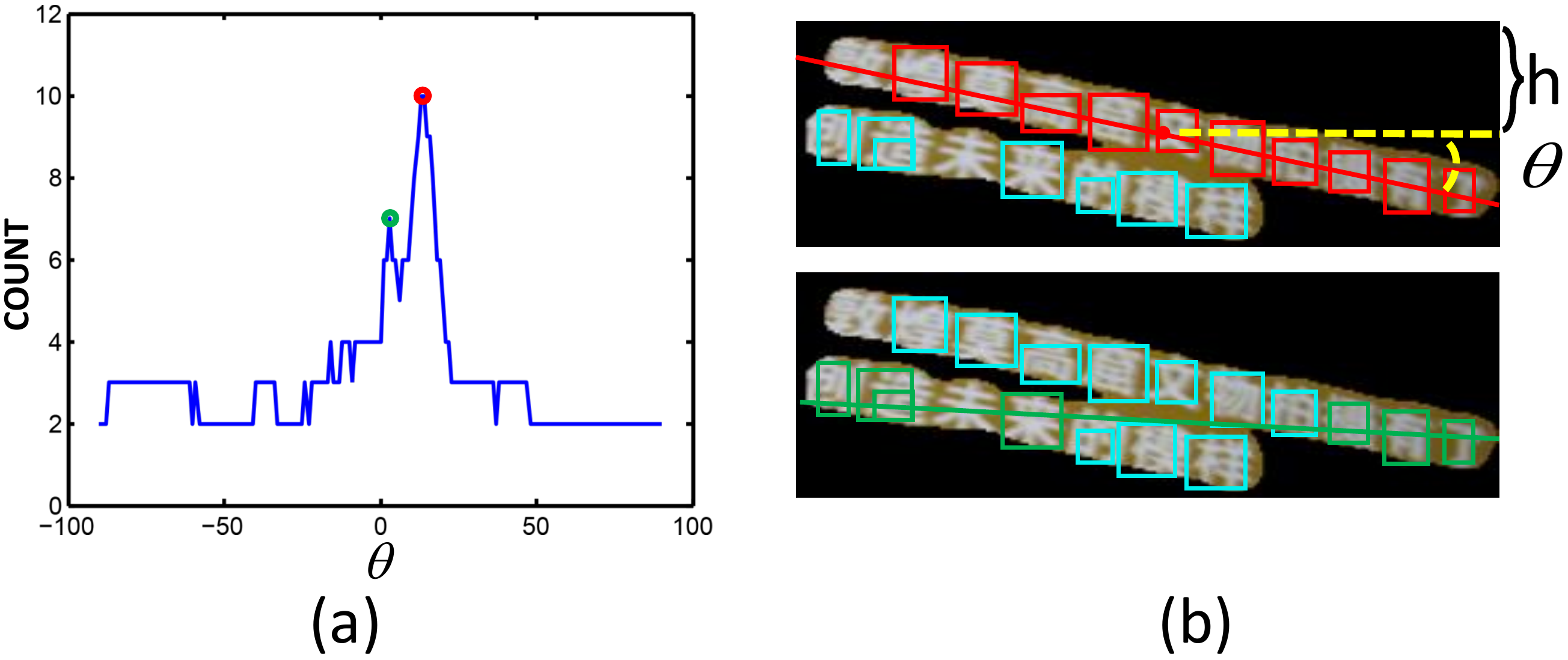}
\vspace{-2mm}
\end{center}
   \caption{(a) The line chart about the counting number of components in different orientations. The right direction has the maximum value (red circle), and the wrong direction has smaller value (green circle); (b) The red line and green line correspond to circles on the line chart.}
\label{fig:orient_estimation}
\vspace{-2mm}
\end{figure}

\subsection{Multi-Oriented Text Line Candidate Generation}\label{Sec:textlineGeneration}

In this section, we introduce how to form multi-oriented text line candidates based on text blocks. Although the text blocks detected by the Text-Block FCN provide coarse localizations of text lines, they are still far from satisfactory. To further extract accurate bounding boxes of text lines, taking the information about the orientation and scale of text into account is required. Character components within a text line or word reveal the scale of text. Besides, the orientation of text can be estimated by analyzing the layout of the character components. At first, we extract the character components within the text blocks by MSER~\cite{Ref:NeumannM12MSER}. Then, similar to many skew correction approaches in document analysis~\cite{postl1986skew}, the orientation of the text lines within a text block is estimated by component projection. Finally, text line candidates are extracted by a novel method that effectively combines block-level (global) cue and component-level (local) cue.

\textbf{Character Components Extraction.}
Our approach uses MSER~\cite{Ref:NeumannM12MSER} to extract the character components (Fig.~\ref{fig:introduction_pipeline}(d)) since MSER is insensitive to variations in scales, orientations, positions, languages and fonts. Two constraints are adopted to remove the most of false components: area and aspect ratio. Specifically, the minimal area ratio of a character candidate needs to be more than the threshold $T_{1}$, and the aspect ratio of them must be limited to $[\frac{1}{T_{2}}, T_{2}]$. Under these two constraints, the most of the false components are excluded.

\textbf{Orientation Estimation.}
In this paper, we assume that text lines from the same text block have a roughly uniform spatial layout, and characters from one text line are in the arrangement of straight or near-straight line. Inspired by projection profile based skew estimation algorithms in documents analysis~\cite{postl1986skew}, we propose a projection method according to counting components, in order to estimate the possible orientation of text lines. Suppose the orientation of text lines within a text block is $\theta$, and the vertical-coordinate offset is $h$, we can draw a line across the text block (as the green or red line is shown in Fig.~\ref{fig:orient_estimation}(b)). And the value of counting components $\Phi(\theta,h)$ equals the number of the character components that are passed through by the line. Since the component number in the right direction often has the maximum value, the possible orientation $\theta_{r}$ can be easily found if we have statistics on the peak value of counting component in all directions (Fig.~\ref{fig:orient_estimation}(a)). By this means, $\theta_{r}$ can be easily calculated as the following formulation:
\begin{equation}
\theta_{r} = \mathop{\arg \max}_{\theta} \mathop{\max}_{h} \Phi(\theta,h)
\end{equation} 
where $\Phi(\theta, h)$ represents the number of components when the orientation is $\theta$ and the vertical-coordinate offset is $h$.

\textbf{Text Line Candidate Generation.}
Different from component based methods~\cite{Ref:NeumannM12MSER,Ref:Epshtein2010,Ref:Huang2014}, the process of generating text line candidates in our approach does not require to catch all the characters within a text line, under the guidance of a text block. First, we divide the components into groups. A pair of the components ($A$ and $B$) within the text block $\alpha$ are grouped together if they satisfy following conditions:
\begin{equation}
\frac{2}{3} < \frac{H(A)}{H(B)} < \frac{3}{2},
\end{equation}
\begin{equation}
-\frac{\pi}{12} < O(A,B) - \theta_{r}(\alpha) < \frac{\pi}{12},
\end{equation}
where $H(A)$ and $H(B)$ represent the heights of $A$ and $B$, $O(A,B)$ represents the orientation of the pair, and $\theta_{r}(\alpha)$ is the estimated orientation of $\alpha$.

Then, for one group $\beta = \{c_i\}$, $c_i$ is $i$-th component, we draw a line $l$ along the orientation $\theta_{r}(\alpha)$ passing the center of $\beta$. The point set $\mathcal{P}$ is defined as:
\begin{equation}
\mathcal{P} = \{p_i\}, p_i \in l \cap \mathbb{B}(\alpha), 
\end{equation}
where $\mathbb{B}(\alpha)$ represents the boundary points of $\alpha$. 

Finally, the minimum bounding box $bb$ of $\beta$ is computed as a text line candidate:
\begin{equation}
bb= \bigcup\{p_1, p_2, ... p_i, c_1, c_2, ..., c_j\}, p_i \in \mathcal{P}, c_j \in \beta,
\end{equation}
where $\bigcup$ denotes the minimum bounding box that contains all points and components. 

\begin{figure}[!h]
\begin{center}
\includegraphics[width=0.9\linewidth]{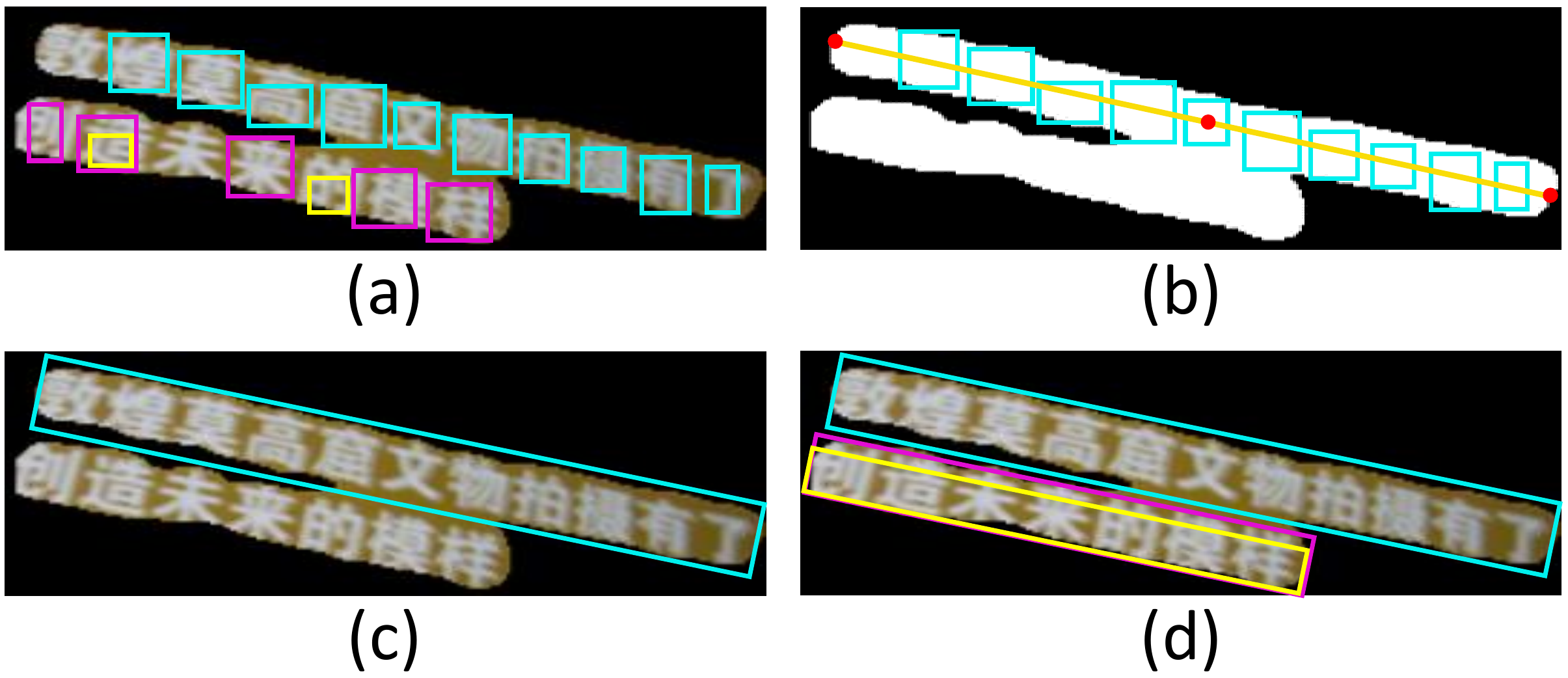}
\end{center}
\vspace{-2mm}
   \caption{The illustration of text line candidate generation. (a) The character components within a text block are divided into groups; (b) The middle red point is the center of a group and the other two red points belong to $\mathcal{P}$; (c) The minimum bounding box is computed as a text line candidate; (d) All the text line candidates of this text block are extracted.}
\label{fig:candidate_found}
\vspace{-2mm}
\end{figure}

Fig.~\ref{fig:candidate_found} illustrates this procedure. We repeat this procedure for each text block to obtain all the text line candidates within an image.
By considering both of the two level cues at the same time, our approach has two advantages compared to component based methods~\cite{Ref:NeumannM12MSER,Ref:Epshtein2010,Ref:Huang2014}. First, under the guidance of text blocks, MSER components are not required to catch all characters accurately. Even though some characters are missed or partially detected by MSER,  the generation of text line candidates will not be affected (such as the three candidates found at Fig.~\ref{fig:candidate_found}). Second, the previous works~\cite{Ref:Yao2012,Ref:Kang2014,yinmulti} of multi-oriented text detection in natural scenes usually estimate the orientation of text based on the character level with some fragile clustering/grouping algorithms. This kind of methods is sensitive to missing characters and non-text noise. Our method estimates the orientation from the holistic profile by using a projection method, which is more efficient and robust than character clustering/grouping based methods.

\subsection{Text Line Candidates Classification}\label{Sec:candidateVerification}
A fraction of the candidates generated in the last stage (Sec.~\ref{Sec:textlineGeneration}) are non-text or redundancy. In order to remove false candidates, we propose two criteria based on the character centroids of text line candidates. To predict the character centroids, we employ another FCN model, named Character-Centroid FCN.


\textbf{Character-Centroid FCN} The Character-Centroid FCN is inherited from the Text-Block FCN (Sec.~\ref{Sec:textBlock}), but only the first 3 convolutional stages are used. Same as the Text-Block FCN, each stage is followed by a $1\times 1$ convolutional layer and a upsampling layer. The fully-connected layers are also replaced with a $1\times 1$ convolutional layer and a sigmoid layer. This network is trained with the cross-entropy loss function as well. In general, the Character-Centroid FCN is a small version of the Text-Block FCN.

\begin{figure}[!h]
\begin{center}
\includegraphics[width=0.92\linewidth]{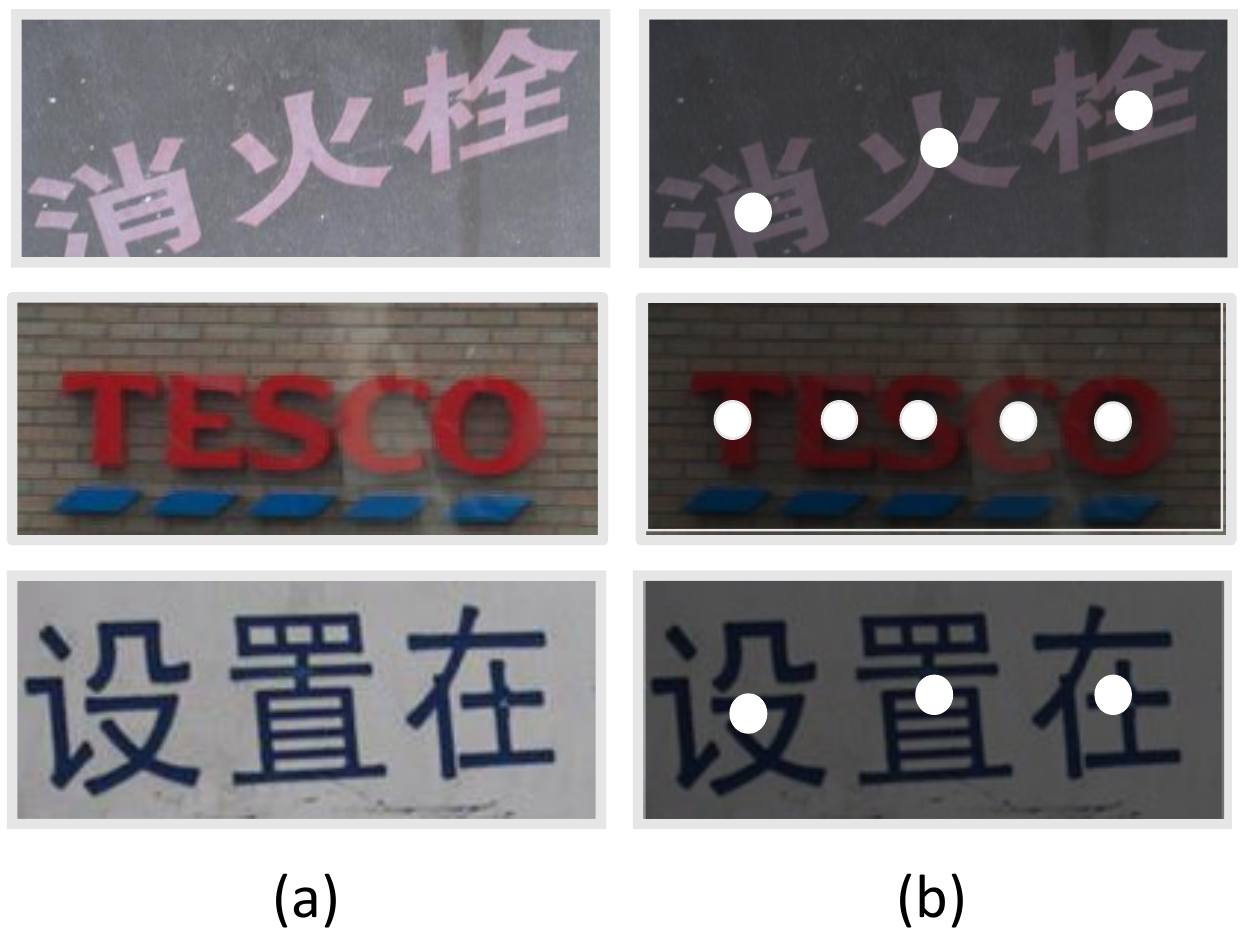}
\end{center}
\vspace{-2mm}
   \caption{The illustration of the ground truth map used for training the Character-Centroid FCN. (a) Input images; (b) The ground truth maps. The white circles in (b) indicate the centroid of characters within the input images.}
\label{fig:Character-level-classifier-ground-truth}
\end{figure}

\begin{figure}[!h]
\begin{center}
\includegraphics[width=0.95\linewidth]{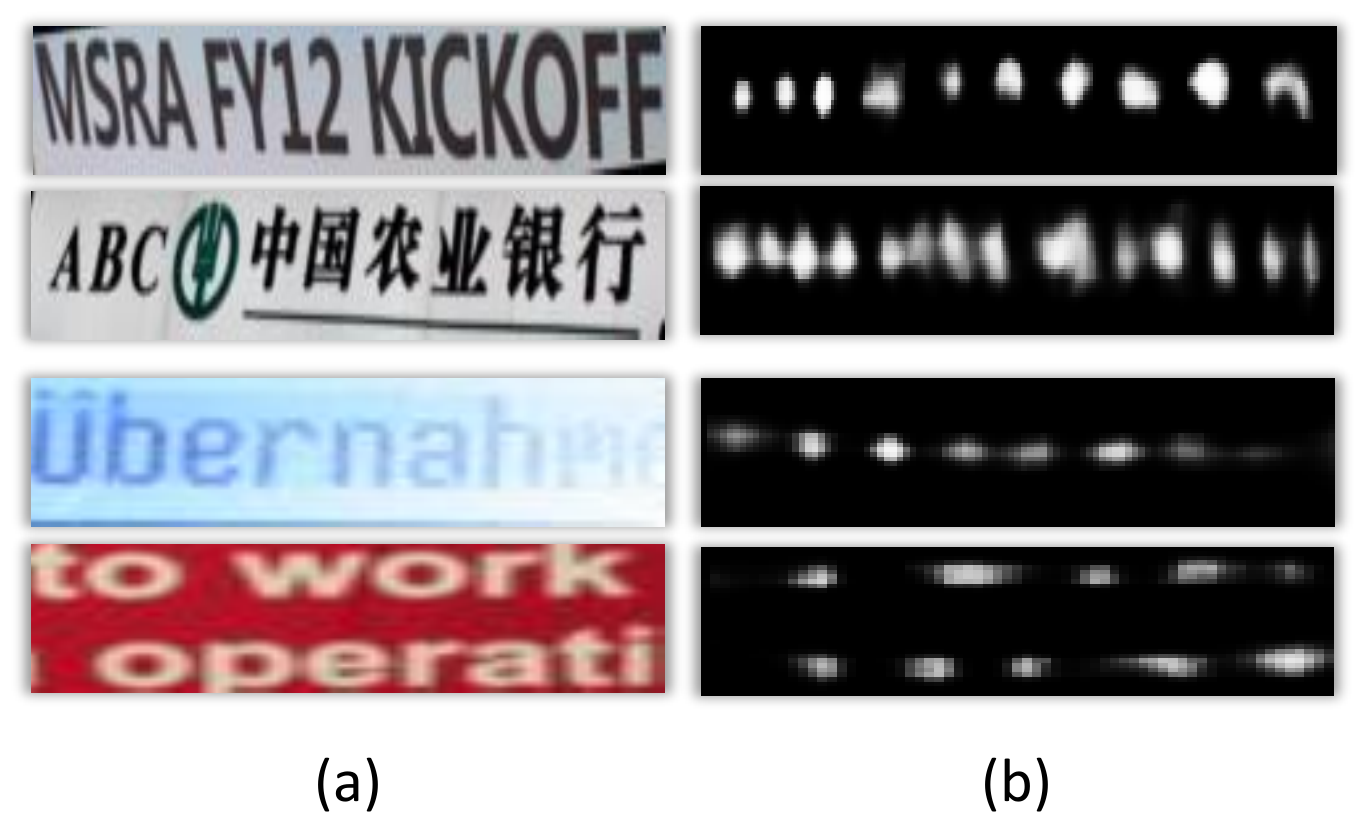}
\end{center}
\vspace{-2mm}
   \caption{Examples of probability maps predicted by the Character-Centroid FCN. (a) Input images; (b) The probability maps of character centroids.}
\label{fig:character_detector}
\end{figure}

Several examples along with ground truth maps are shown in Fig.~\ref{fig:Character-level-classifier-ground-truth}. The positive region of the ground truth map consists of the pixels whose distance to the character centroids is less than $15\%$ of the height of the corresponding character. In the testing phase, we can obtain the centroid probability map of a text line candidate at first. Then, extreme points $\mathcal{E} = \{(e_i, s_i)\}$ on the map are collected as the centroids, where $e_i$ represents $i$-th extreme point, and $s_i$ represents the score defined as the value of the probability map on $e_i$. Several examples are shown in Fig.~\ref{fig:character_detector}. 

In order to remove false candidates, two intuitive yet effective criteria based on intensity and geometric properties are adopted, after the centroids are obtained:

\textbf{Intensity criterion}. For a text line candidate, if the number of the character centroids $n_c < 2$, or the average score of the centroids $s_{avg} < 0.6$, we regard it as a false text line candidate. The average score of the centroids is defined as:
\begin{equation}
s_{avg} = \frac{1}{n_c} \sum_{i = 1}^{n_c} s_i,
\end{equation}

\textbf{Geometric criterion}. The arrangement of the characters within a text line candidate is always approximated to a straight line. We adopt the mean of orientation angles $\mu$ and the standard deviation $\sigma$ of orientation angles between the centroids to characterize these properties. $\mu$ and $\sigma$ are defined as:
\begin{equation}
\mu = \frac{1}{n_c} \sum_{i = 1}^{n_c} \sum_{j = 1}^{n_c} O(e_i, e_j),
\end{equation}
\begin{equation}
\sigma = \sqrt{\frac{1}{n_c} \sum_{i=1}^{n_c} \sum_{j=1}^{n_c}(O(e_i, e_j) - \mu)^2},
\end{equation}
where $O(e_i, e_j)$ denotes the orientation angle between $e_i$ and $e_j$. In practice, we only reserve the candidates whose $\mu < \frac{\pi}{32}$ and $\sigma < \frac{\pi}{16}$.

Through the above two constraints, the false text line candidates are excluded, but there are still some redundant candidates. To further remove the redundant candidates, a standard non-maximum suppression is applied to remaining candidates, and the score that used in non-maximum suppression is defined as the sum of the score of all the centroids.

\begin{figure*}[ht]
\begin{center}
\includegraphics[width=0.85\linewidth]{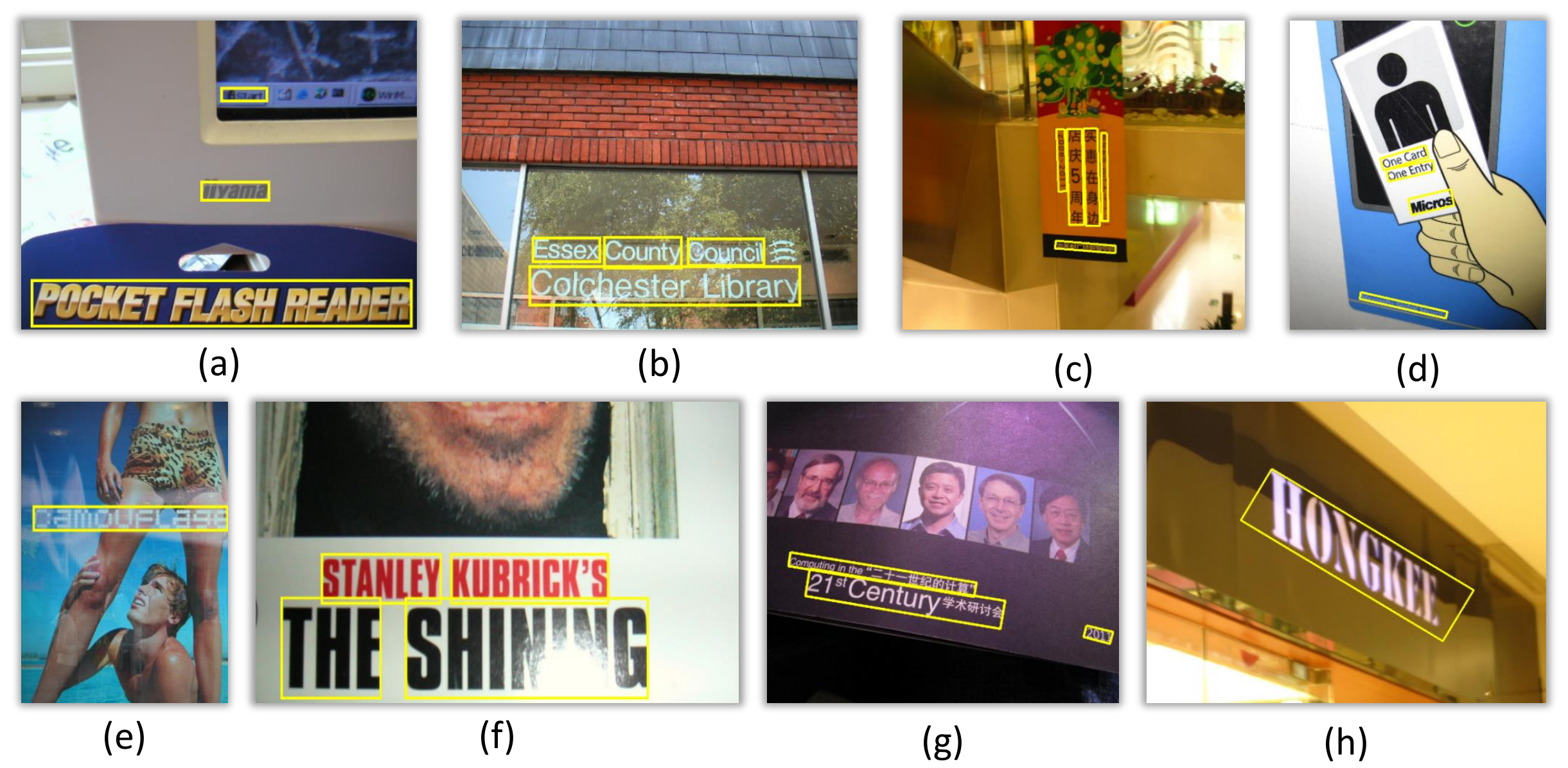}
\end{center}
\vspace{-2mm}
   \caption{Detection examples of the proposed method on MSRA-TD500 and ICDAR2013.}
\label{fig:DetectionExamples}
\end{figure*}

\begin{figure}[!h]
\begin{center}
\includegraphics[width=0.95\linewidth]{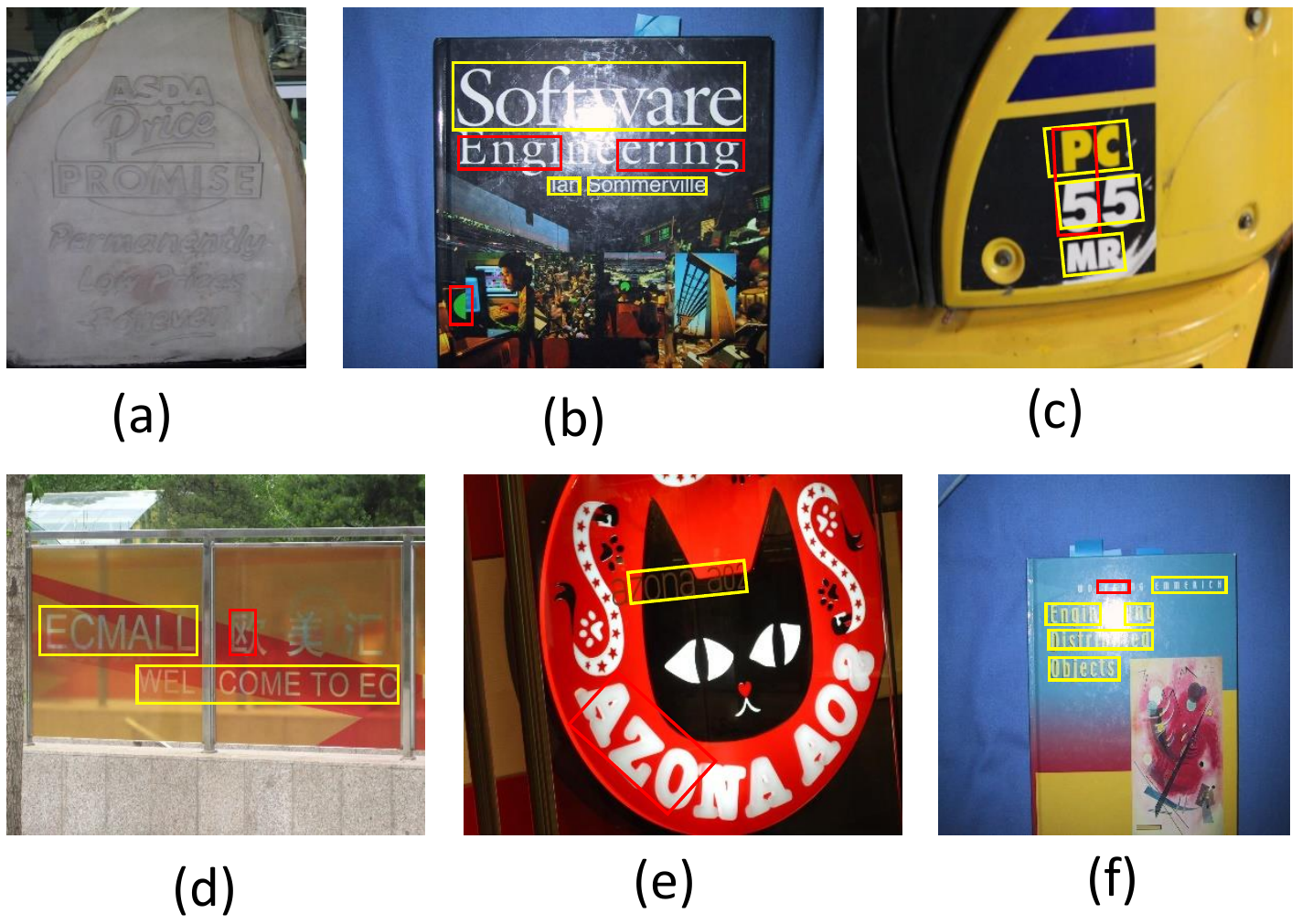}
\end{center}
\vspace{-2mm}
   \caption{Several failure cases of the proposed method.}
\label{fig:failedCases}
\end{figure}

\section{Experiments} \label{Sec:Experiments}

To fully compare the proposed method with competing methods, we evaluate our method on several recent standard benchmarks: ICDAR2013, ICDAR2015 and MSRA-TD500. 


\subsection{Datasets}\label{Sec:Datasets}
We evaluate our method on three datasets, where the first two are multi-oriented text datasets, and the third one is a horizontal text dataset.

\noindent\textbf{MSRA-TD500}. The MSRA-TD500 dataset introduced in~\cite{Ref:Yao2012}, is a multi-orientation text dataset including 300 training images and 200 testing images. The dataset contains text in two languages, namely Chinese and English. This dataset is very challenging due to the large variation in fonts, scales, colors and orientations. Here, we followed the evaluation protocol employed by~\cite{Ref:Yao2012}, which considers both of the area overlap ratios and the orientation differences between predictions and the ground truth.

\noindent\textbf{ICDAR2015 - Incidental Scene Text dataset}. The ICDAR2015 - Incidental Scene Text dataset is the benchmark of ICDAR2015 Incidental Scene Text Competition. This dataset includes 1000 training images and 500 testing images. Different from the previous ICDAR competition, in which the text are well-captured, horizontal, and typically centered in images, these datasets focus on the incidental scene where text may appear in any orientation and any location with small size or low resolution. The evaluation protocol of this dataset inherits from ~\cite{Ref:Lucas2005}. Note that this competition provides an online evaluation system and our method is evaluated in the same way. 

Unlike MSRA-TD500, in which the ground truth is marked at the sentence level, the annotations of ICDAR2015 are word level. To satisfy the requirement of ICDAR 2015 measurement, we perform the word partition on the text lines generated by our method according to the blanks between words.

\noindent\textbf{ICDAR2013}. The ICDAR 2013 dataset is a horizontal text database which is used in previous ICDAR competitions. This dataset consists of 229 images for training and 233 images for testing. The evaluation algorithm is introduced by \cite{Ref:Karatzas2013} and we evaluate our method on the ICDAR2013 online evaluation system. Since this dataset also provides word-level annotations, we adopt the same word partition procedure as we did on ICDAR 2015 dataset.

\subsection{Implementation Details} \label{Sec:ImplementionDetails}
In the proposed method, two models are used: the Text-Block FCN is used to generate text salient maps and the Character-Centroid FCN is used to predict the centroids of characters. Both of the two models are trained under the same network configuration. Similar to ~\cite{long2015fully,xie2015holistically}, we also use fine-tuning with the pre-trained VGG-16 network. The two models both are trained $20 \times 10^{5}$ iterations in all. Learning rates start from $10^{-6}$, and are multiplied by $\frac{1}{10}$ after $10 \times 10^{5}$ and $15 \times 10^{5}$ iterations. Weight decays are $0.0001$, and momentums are $0.9$. No dropout or batch normalization is used in our model.

All training images are harvested from the training set of ICDAR2013, ICDAR2015 and MSRA-TD500 with data augmentation. In the training phase of the Text-Block FCN, we randomly crop $30K$ $500\times500$ patches from the images as training examples. To compute the salient map in the testing phase, each image is proportionally resized to three scales, where the heights are $200$, $500$ and $1000$ pixels respectively. For the Character-Centroid FCN, the patches ($32\times256$ pixels) around the word level ground truth are collected as training examples. We randomly collect $100K$ patches in the training phase. In the testing phase, we rotate text line candidates to horizontal orientation and proportionally resize them to $32$ pixels height. For all experiments, threshold values are: $T_1 = 0.5\%$, $T_2 = 3$.

The proposed method is implemented with Torch7 and Matlab (with C/C++ mex functions) and runs on a workstation(2.0GHz 8-core CPU, 64G RAM, GTX TitanX and Windows 64-bit OS) for all the experiments.

\subsection{Experimental Results} \label{Sec:ExperimentalResults}
\noindent\textbf{MSRA-TD500}. As shown in Tab.~\ref{Tab:MSRADetection}, our method outperforms other methods in both precision and recall on MSRA-TD500. The proposed method achieves precision $0.83$, recall $0.67$ and f-measure $0.74$. Compared to~\cite{yinmulti}, our method obtains significant improvements on precision ($0.02$), recall ($0.04$) and f-measure ($0.03$). In addition, the time cost of our method is reported in Tab.~\ref{Tab:MSRADetection}. Benefiting from GPU acceleration, our method takes $2.1$s for each image in average on MSRA-TD500.

\begin{table}
\caption{Performance comparisons on the MSRA-TD500 dataset.}
\label{Tab:MSRADetection}
\begin{center}
\begin{tabular}{|c|c|c|c|c|}
\hline
\textbf{\begin{footnotesize}Algorithm\end{footnotesize}}&\textbf{\begin{footnotesize}Precision\end{footnotesize}}&\textbf{\begin{footnotesize}Recall\end{footnotesize}}&\textbf{\begin{footnotesize}F-measure\end{footnotesize}}&\textbf{\begin{footnotesize}Time cost\end{footnotesize}}   \\
\hline
\hline
\begin{footnotesize}
Proposed
\end{footnotesize}                  &
\begin{footnotesize}
\textbf{0.83}
\end{footnotesize} &
\begin{footnotesize}
\textbf{0.67}
\end{footnotesize} &
\begin{footnotesize}
\textbf{0.74}
\end{footnotesize} &
\begin{footnotesize}
2.1s
\end{footnotesize}\\
\hline
\begin{footnotesize}
Yin~\emph{et al.}~\cite{yinmulti}
\end{footnotesize}                  &
\begin{footnotesize}
0.81
\end{footnotesize} &
\begin{footnotesize}
0.63
\end{footnotesize} &
\begin{footnotesize}
0.71
\end{footnotesize} &
\begin{footnotesize}
1.4s
\end{footnotesize}\\
\hline
\begin{footnotesize}
Kang~\emph{et al.}~\cite{Ref:Kang2014}
\end{footnotesize}                  &
\begin{footnotesize}
0.71
\end{footnotesize} &
\begin{footnotesize}
0.62
\end{footnotesize} &
\begin{footnotesize}
0.66
\end{footnotesize} &
\begin{footnotesize}
-
\end{footnotesize}\\
\hline
\begin{footnotesize}
Yin~\emph{et al.}~\cite{Ref:Yin2014}
\end{footnotesize}                  &
\begin{footnotesize}
0.71
\end{footnotesize} &
\begin{footnotesize}
0.61
\end{footnotesize} &
\begin{footnotesize}
0.65
\end{footnotesize} &
\begin{footnotesize}
\textbf{0.8s}
\end{footnotesize}  \\
\hline
\begin{footnotesize}
Yao~\emph{et al.}~\cite{Ref:Yao2012}
\end{footnotesize}                  &
\begin{footnotesize}
0.63
\end{footnotesize} &
\begin{footnotesize}
0.63
\end{footnotesize} &
\begin{footnotesize}
0.60
\end{footnotesize} &
\begin{footnotesize}
7.2s
\end{footnotesize}\\
\hline
\end{tabular}
\end{center}
\vspace{-2mm}
\end{table}

\noindent\textbf{ICADR2015 - Incidental Scene Text}. As this dataset has been released recently for the competition in ICDAR2015, there is no literature to report the experimental result on it. Therefore, we collect competition results~\cite{karatzasicdar} as listed in Tab.~\ref{Tab:ICDAR2015Detection} for comprehensive comparisons. Our method achieves the best F-measure over all methods. 

\begin{table}
\caption{Performance of different algorithms evaluated on the ICDAR2015 dataset. The comparison results are collected from ICDAR 2015 Competition on Robust Reading~\cite{karatzasicdar}.}
\label{Tab:ICDAR2015Detection}
\begin{center}
\begin{tabular}{|c|c|c|c|}
\hline
\textbf{\begin{footnotesize}Algorithm\end{footnotesize}}&\textbf{\begin{footnotesize}Precision\end{footnotesize}}&\textbf{\begin{footnotesize}Recall\end{footnotesize}}&\textbf{\begin{footnotesize}F-measure\end{footnotesize}}   \\
\hline
\hline
\begin{footnotesize}
Proposed
\end{footnotesize}                  &
\begin{footnotesize}
0.71
\end{footnotesize} &
\begin{footnotesize}
0.43
\end{footnotesize} &
\begin{footnotesize}
\textbf{0.54}
\end{footnotesize} \\
\hline
\begin{footnotesize}
StradVision-2
\end{footnotesize}                  &
\begin{footnotesize}
\textbf{0.77}
\end{footnotesize} &
\begin{footnotesize}
0.37
\end{footnotesize} &
\begin{footnotesize}
0.50
\end{footnotesize} \\
\hline
\begin{footnotesize}
StradVision-1
\end{footnotesize}                  &
\begin{footnotesize}
0.53
\end{footnotesize} &
\begin{footnotesize}
0.46
\end{footnotesize} &
\begin{footnotesize}
0.50
\end{footnotesize} \\
\hline
\begin{footnotesize}
NJU\_Text
\end{footnotesize}                  &
\begin{footnotesize}
0.70
\end{footnotesize} &
\begin{footnotesize}
0.36
\end{footnotesize} &
\begin{footnotesize}
0.47
\end{footnotesize} \\
\hline
\begin{footnotesize}
AJOU
\end{footnotesize}                  &
\begin{footnotesize}
0.47
\end{footnotesize} &
\begin{footnotesize}
\textbf{0.47}
\end{footnotesize} &
\begin{footnotesize}
0.47
\end{footnotesize} \\
\hline
\begin{footnotesize}
HUST\_MCLAB
\end{footnotesize}                  &
\begin{footnotesize}
0.44
\end{footnotesize} &
\begin{footnotesize}
0.38
\end{footnotesize} &
\begin{footnotesize}
0.41
\end{footnotesize} \\
\hline
\begin{footnotesize}
Deep2Text-MO
\end{footnotesize}                  &
\begin{footnotesize}
0.50
\end{footnotesize} &
\begin{footnotesize}
0.32
\end{footnotesize} &
\begin{footnotesize}
0.39
\end{footnotesize} \\
\hline
\begin{footnotesize}
CNN Proposal
\end{footnotesize}                  &
\begin{footnotesize}
0.35
\end{footnotesize} &
\begin{footnotesize}
0.34
\end{footnotesize} &
\begin{footnotesize}
0.35
\end{footnotesize} \\
\hline
\begin{footnotesize}
TextCatcher-2
\end{footnotesize}                  &
\begin{footnotesize}
0.25
\end{footnotesize} &
\begin{footnotesize}
0.34
\end{footnotesize} &
\begin{footnotesize}
0.29
\end{footnotesize} \\
\hline
\end{tabular}
\end{center}
\vspace{-2mm}
\end{table}

\noindent\textbf{ICDAR 2013}. We also test our method on the ICDAR2013 dataset, which is the most popular for horizontal text detection. As shown in Tab.~\ref{Tab:ICDAR2013Detection}, the proposed method achieves $0.88$, $0.78$, $0.83$ in precision, recall and F-measure, respectively, outperforming all other recent methods only designed for horizontal text.  

\begin{table}
\caption{Performance of different algorithms evaluated on the ICDAR 2013 dataset.}
\label{Tab:ICDAR2013Detection}
\begin{center}
\begin{tabular}{|c|c|c|c|}
\hline
\textbf{\begin{footnotesize}Algorithm\end{footnotesize}}&\textbf{\begin{footnotesize}Precision\end{footnotesize}}&\textbf{\begin{footnotesize}Recall\end{footnotesize}}&\textbf{\begin{footnotesize}F-measure\end{footnotesize}}   \\
\hline
\hline
\begin{footnotesize}
Proposed
\end{footnotesize}                  &
\begin{footnotesize}
0.88
\end{footnotesize} &
\begin{footnotesize}
\textbf{0.78}
\end{footnotesize} &
\begin{footnotesize}
\textbf{0.83}
\end{footnotesize} \\
\hline
\begin{footnotesize}
Zhang~\emph{et al.}~\cite{zhang2015symmetry}
\end{footnotesize}                  &
\begin{footnotesize}
0.88
\end{footnotesize} &
\begin{footnotesize}
0.74
\end{footnotesize} &
\begin{footnotesize}
0.80
\end{footnotesize} \\
\hline
\begin{footnotesize}
Tian~\emph{et al.}~\cite{tiantext}
\end{footnotesize}                  &
\begin{footnotesize}
0.85
\end{footnotesize} &
\begin{footnotesize}
0.76
\end{footnotesize} &
\begin{footnotesize}
0.80
\end{footnotesize} \\
\hline
\begin{footnotesize}
Lu~\emph{et al.}~\cite{lu2015scene}
\end{footnotesize}                  &
\begin{footnotesize}
\textbf{0.89}
\end{footnotesize} &
\begin{footnotesize}
0.70
\end{footnotesize} &
\begin{footnotesize}
0.78
\end{footnotesize} \\
\hline
\begin{footnotesize}
iwrr2014~\cite{Ref:Zamberletti2014}
\end{footnotesize}                  &
\begin{footnotesize}
0.86
\end{footnotesize} &
\begin{footnotesize}
0.70
\end{footnotesize} &
\begin{footnotesize}
0.77
\end{footnotesize} \\
\hline
\begin{footnotesize}
USTB TexStar~\cite{Ref:Yin2014}
\end{footnotesize}                  &
\begin{footnotesize}
0.88
\end{footnotesize} &
\begin{footnotesize}
0.66
\end{footnotesize} &
\begin{footnotesize}
0.76
\end{footnotesize} \\
\hline
\begin{footnotesize}
Text Spotter~\cite{Ref:NeumannM12MSER}
\end{footnotesize}                  &
\begin{footnotesize}
0.88
\end{footnotesize} &
\begin{footnotesize}
0.65
\end{footnotesize} &
\begin{footnotesize}
0.74
\end{footnotesize} \\
\hline
\begin{footnotesize}
Yin~\emph{et al.}~\cite{yinmulti}
\end{footnotesize}                  &
\begin{footnotesize}
0.84
\end{footnotesize} &
\begin{footnotesize}
0.65
\end{footnotesize} &
\begin{footnotesize}
0.73
\end{footnotesize} \\
\hline
\begin{footnotesize}
CASIA\_NLPR~\cite{Ref:ICDAR2013}
\end{footnotesize}                  &
\begin{footnotesize}
0.79
\end{footnotesize} &
\begin{footnotesize}
0.68
\end{footnotesize} &
\begin{footnotesize}
0.73
\end{footnotesize} \\
\hline
\begin{footnotesize}
Text\_Detector\_CASIA~\cite{Ref:Shi2013B}
\end{footnotesize}                  &
\begin{footnotesize}
0.85
\end{footnotesize} &
\begin{footnotesize}
0.63
\end{footnotesize} &
\begin{footnotesize}
0.72
\end{footnotesize} \\
\hline
\begin{footnotesize}
I2R\_NUS\_FAR~\cite{Ref:ICDAR2013}
\end{footnotesize}                  &
\begin{footnotesize}
0.75
\end{footnotesize} &
\begin{footnotesize}
0.69
\end{footnotesize} &
\begin{footnotesize}
0.72
\end{footnotesize} \\
\hline
\begin{footnotesize}
I2R\_NUS~\cite{Ref:ICDAR2013}
\end{footnotesize}                  &
\begin{footnotesize}
0.73
\end{footnotesize} &
\begin{footnotesize}
0.66
\end{footnotesize} &
\begin{footnotesize}
0.69
\end{footnotesize} \\
\hline
\begin{footnotesize}
TH-TextLoc~\cite{Ref:ICDAR2013}
\end{footnotesize}                  &
\begin{footnotesize}
0.70
\end{footnotesize} &
\begin{footnotesize}
0.65
\end{footnotesize} &
\begin{footnotesize}
0.67
\end{footnotesize} \\
\hline
\end{tabular}
\end{center}
\vspace{-7mm}
\end{table}

The consistent top performance achieved on the three datasets demonstrates the effectiveness and generality of the proposed method. Besides the quantitative experimental results, several detection examples under various challenging cases of the proposed method on the MSRA-TD500 and ICDAR2013 datasets are shown in Fig.~\ref{fig:DetectionExamples}. As can be seen, our method successfully detects the text with inner texture in Fig.~\ref{fig:DetectionExamples} (a), non-uniform illumination in Fig.~\ref{fig:DetectionExamples} (b) (f), dot fonts (Fig.~\ref{fig:DetectionExamples} (e)), broken strokes (Fig.~\ref{fig:DetectionExamples} (h)), multiple orientations (Fig.~\ref{fig:DetectionExamples} (c), and (d)), perspective distortion (Fig.~\ref{fig:DetectionExamples} (h)) and mixture of multi-language (Fig.~\ref{fig:DetectionExamples} (g)).

\subsection{Impact of Parameters} \label{Sec:parameters}
\begin{figure}[!h]
\begin{center}
\includegraphics[width=0.95 \linewidth]{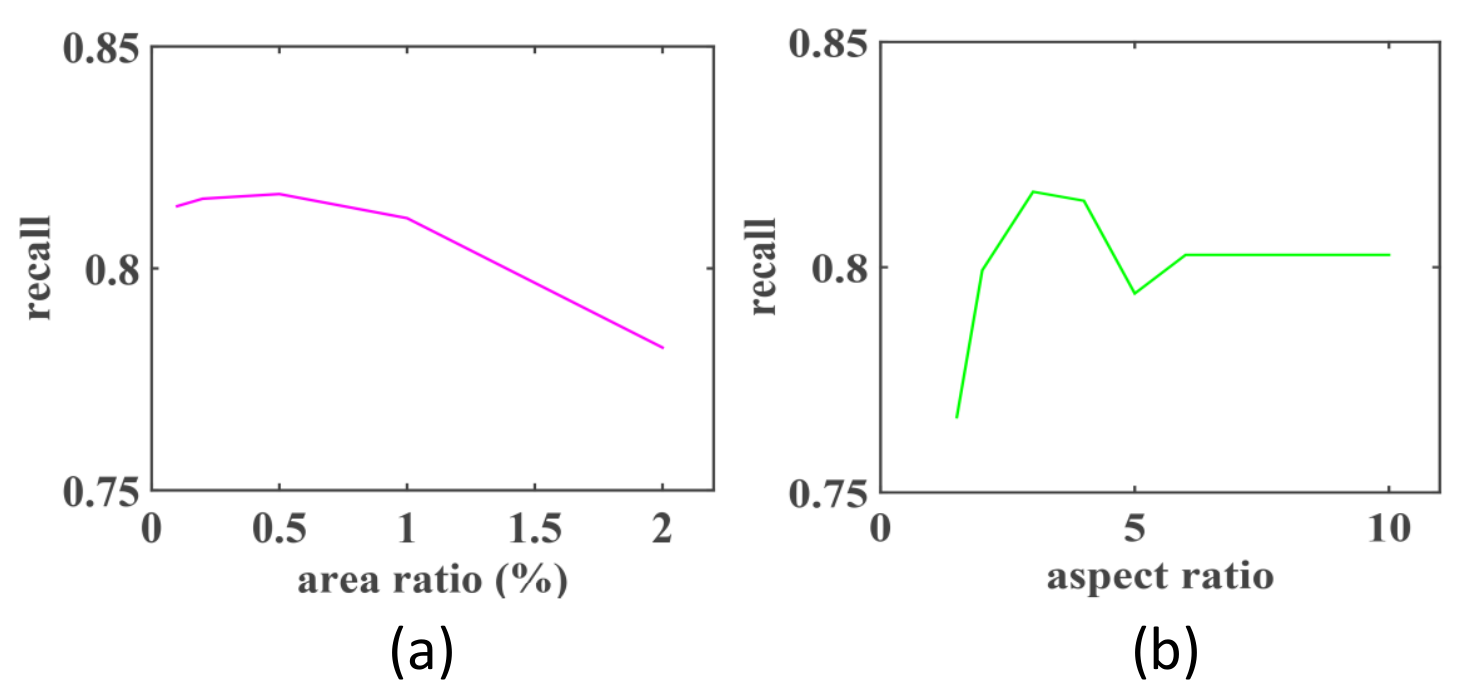}
\end{center}
\vspace{-5mm}
	\caption{The recall of text line candidates with different $T_1$ and $T_2$.}
\label{fig:ImpactParameters}
\end{figure}

In this section, we investigate the effect of parameters $T_1$ and $T_2$, which are used to extract MSER components for computing text line candidates. The performance of different parameters is computed on MSRA-TD500. Fig.~\ref{fig:ImpactParameters} (a) and Fig.~\ref{fig:ImpactParameters} (b) show how the recall of text line candidates changes under the different settings of $T_1$ and $T_2$. As we can see, the recall of text line candidates is insensitive to the change of $T_1$ and $T_2$ in a large range. This proves our method does not depend on the quality of character candidates.

\subsection{Limitations of the Proposed Algorithm} \label{Sec:limitations}
The proposed method achieves excellent performance and is able to deal with several challenging cases. However, our method still has a great gap to achieve a perfect performance. Several failure cases are illustrated in Fig.~\ref{fig:failedCases}. As can be seen, false positives and missing characters may appear in certain situations, such as extremely low contrast (Fig.~\ref{fig:failedCases} (a)), curvature (Fig.~\ref{fig:failedCases} (e)), strong reflect light (Fig.~\ref{fig:failedCases} (b) (f)), too closed text lines (Fig.~\ref{fig:failedCases} (c)), or tremendous gap between characters (Fig.~\ref{fig:failedCases} (d)). Another limitation is the speed of the proposed method, which is still far from the requirement of real-time systems.

\section{Conclusion} \label{Sec:Conclusion}
In this paper, we presented a novel framework for multi-oriented scene text detection. The main idea that integrates semantic labeling by FCN and MSER provides a natural solution for handling multi-oriented text. The superior performance over other competing methods in the literature on both horizontal and multi-oriented text detection benchmarks verifies that combining local and global cues for text line localization is an interesting direction that is worthy of being studied. In the future, we could extend the proposed method to an end-to-end text recognition system. 

\section*{Acknowledgement}
This work was primarily supported by National Natural Science Foundation of China (NSFC) (No. 61222308, No. 61573160, No. 61572207, and No. 61303095), and Open Project Program of the State Key Laboratory of Digital Publishing Technology (No. F2016001).

{\small
\bibliographystyle{ieee}
\bibliography{cvpr-textdet-fcn}

\begin{thebibliography}{10}\itemsep=-1pt

\bibitem{Ref:ICDAR2013}
{ICDAR} 2013 robust reading competition challenge 2 results.
\newblock \url{http://dag.cvc.uab.es/icdar2013competition}, 2014.
\newblock [Online; accessed 11-November-2014].

\bibitem{Ref:Bissacco2013}
A.~Bissacco, M.~Cummins, Y.~Netzer, and H.~Neven.
\newblock {PhotoOCR}: Reading text in uncontrolled conditions.
\newblock In {\em Proc. of ICCV}, 2013.

\bibitem{Ref:Chen2004}
X.~Chen and A.~Yuille.
\newblock Detecting and reading text in natural scenes.
\newblock In {\em Proc. of CVPR}, 2004.

\bibitem{Ref:Epshtein2010}
B.~Epshtein, E.~Ofek, and Y.~Wexler.
\newblock Detecting text in natural scenes with stroke width transform.
\newblock In {\em Proc. of CVPR}, 2010.

\bibitem{Ref:Huang2013}
W.~Huang, Z.~Lin, J.~Yang, and J.~Wang.
\newblock Text localization in natural images using stroke feature transform
  and text covariance descriptors.
\newblock In {\em Proc. of ICCV}, 2013.

\bibitem{Ref:Huang2014}
W.~Huang, Y.~Qiao, and X.~Tang.
\newblock Robust scene text detection with convolution neural network induced
  mser trees.
\newblock In {\em Proc. of ECCV}, 2014.

\bibitem{jaderberg2014reading}
M.~Jaderberg, K.~Simonyan, A.~Vedaldi, and A.~Zisserman.
\newblock Reading text in the wild with convolutional neural networks.
\newblock {\em IJCV}, pages 1--20, 2014.

\bibitem{Ref:Jaderberg2014}
M.~Jaderberg, A.~Vedaldi, and A.~Zisserman.
\newblock Deep features for text spotting.
\newblock In {\em Proc. of ECCV}, 2014.

\bibitem{Ref:Kang2014}
L.~Kang, Y.~Li, and D.~Doermann.
\newblock Orientation robust text line detection in natural images.
\newblock In {\em Proc. of CVPR}, 2014.

\bibitem{karatzasicdar}
D.~Karatzas, L.~Gomez-Bigorda, A.~Nicolaou, S.~Ghosh, A.~Bagdanov, M.~Iwamura,
  J.~Matas, L.~Neumann, V.~R. Chandrasekhar, S.~Lu, et~al.
\newblock Icdar 2015 competition on robust reading.
\newblock In {\em Proc. of ICDAR}, 2015.

\bibitem{Ref:Karatzas2013}
D.~Karatzas, F.~Shafait, S.~Uchida, M.~Iwamura, L.~G. i~Bigorda, S.~R. Mestre,
  J.~Mas, D.~F. Mota, J.~A. Almazan, and L.~P. de~las Heras.
\newblock {ICDAR} 2013 robust reading competition.
\newblock In {\em Proc. of ICDAR}, 2013.

\bibitem{long2015fully}
J.~Long, E.~Shelhamer, and T.~Darrell.
\newblock Fully convolutional networks for semantic segmentation.
\newblock In {\em Proc. of CVPR}, 2015.

\bibitem{lu2015scene}
S.~Lu, T.~Chen, S.~Tian, J.-H. Lim, and C.-L. Tan.
\newblock Scene text extraction based on edges and support vector regression.
\newblock {\em IJDAR}, 18(2):125--135, 2015.

\bibitem{Ref:Lucas2005}
S.~M. Lucas.
\newblock {ICDAR} 2005 text locating competition results.
\newblock In {\em Proc. of ICDAR}, 2005.

\bibitem{Ref:Neumann2010}
L.~Neumann and J.~Matas.
\newblock A method for text localization and recognition in real-world images.
\newblock In {\em Proc. of ACCV}, 2010.

\bibitem{Ref:NeumannM12MSER}
L.~Neumann and J.~Matas.
\newblock Real-time scene text localization and recognition.
\newblock In {\em Proc. of CVPR}, 2012.

\bibitem{Ref:Neumann2013B}
L.~Neumann and J.~Matas.
\newblock Scene text localization and recognition with oriented stroke
  detection.
\newblock In {\em Proc. of ICCV}, 2013.

\bibitem{Ref:Pan2011}
Y.~Pan, X.~Hou, and C.~Liu.
\newblock A hybrid approach to detect and localize texts in natural scene
  images.
\newblock {\em IEEE Trans. on Image Processing}, 20(3):800--813, 2011.

\bibitem{postl1986skew}
C.~Postl.
\newblock Detection of linear oblique structures and skew scan in digitized
  documents.
\newblock In {\em Proc. of ICPR}, 1986.

\bibitem{qin2016fast}
S.~Qin and R.~Manduchi.
\newblock A fast and robust text spotter.
\newblock In {\em Proc. of WACV}, 2016.

\bibitem{Ref:Shi2013B}
C.~Shi, C.~Wang, B.~Xiao, Y.~Zhang, and S.~Gao.
\newblock Scene text detection using graph model built upon maximally stable
  extremal regions.
\newblock {\em Pattern Recognition Letters}, 34(2):107--116, 2013.

\bibitem{simonyan2014very}
K.~Simonyan and A.~Zisserman.
\newblock Very deep convolutional networks for large-scale image recognition.
\newblock 2015.

\bibitem{tiantext}
S.~Tian, Y.~Pan, C.~Huang, S.~Lu, K.~Yu, and C.~Lim~Tan.
\newblock Text flow: A unified text detection system in natural scene images.
\newblock In {\em Proc. of CVPR}, 2015.

\bibitem{Ref:Wang2010}
K.~Wang and S.~Belongie.
\newblock Word spotting in the wild.
\newblock In {\em Proc. of ECCV}, 2010.

\bibitem{Ref:Wang2012ICPR}
T.~Wang, D.~J. Wu, A.~Coates, and A.~Y. Ng.
\newblock End-to-end text recognition with convolutional neural networks.
\newblock In {\em Proc. of ICPR}, 2012.

\bibitem{xie2015holistically}
S.~Xie and Z.~Tu.
\newblock Holistically-nested edge detection.
\newblock 2015.

\bibitem{xiongtext}
B.~Xiong and K.~Grauman.
\newblock Text detection in stores using a repetition prior.
\newblock In {\em Proc. of WACV}, 2016.

\bibitem{Ref:Yao2014C}
C.~Yao, X.~Bai, and W.~Liu.
\newblock A unified framework for multi-oriented text detection and
  recognition.
\newblock {\em IEEE Trans. on Image Processing}, 23(11):4737--4749, 2014.

\bibitem{Ref:Yao2012}
C.~Yao, X.~Bai, W.~Liu, Y.~Ma, and Z.~Tu.
\newblock Detecting texts of arbitrary orientations in natural images.
\newblock In {\em Proc. of CVPR}, 2012.

\bibitem{Ref:Ye2014}
Q.~Ye and D.~Doermann.
\newblock Text detection and recognition in imagery: A survey.
\newblock {\em IEEE Trans. on PAMI}, (99), 2014.

\bibitem{Ref:Yi2011}
C.~Yi and Y.~Tian.
\newblock Text string detection from natural scenes by structure-based
  partition and grouping.
\newblock {\em IEEE Trans. on Image Processing}, 20(9):2594--2605, 2011.

\bibitem{yinmulti}
X.-C. Yin, W.-Y. Pei, J.~Zhang, and H.-W. Hao.
\newblock Multi-orientation scene text detection with adaptive clustering.
\newblock {\em IEEE Trans. on PAMI}, (1):1--1.

\bibitem{Ref:Yin2014}
X.~C. Yin, X.~Yin, K.~Huang, and H.~Hao.
\newblock Robust text detection in natural scene images.
\newblock {\em IEEE Trans. on PAMI}, 36(5):970--983, 2014.

\bibitem{Ref:Zamberletti2014}
A.~Zamberletti, L.~Noce, and I.~Gallo.
\newblock Text localization based on fast feature pyramids and multi-resolution
  maximally stable extremal regions.
\newblock In {\em Proc. of ACCV workshop}, 2014.

\bibitem{zhang2015symmetry}
Z.~Zhang, W.~Shen, C.~Yao, and X.~Bai.
\newblock Symmetry-based text line detection in natural scenes.
\newblock In {\em Proc. of CVPR}, 2015.

\bibitem{zhu2016scene}
Y.~Zhu, C.~Yao, and X.~Bai.
\newblock Scene text detection and recognition: Recent advances and future
  trends.
\newblock {\em Frontiers of Computer Science}, 10(1):19--36, 2016.

\end{thebibliography}
}
\end{document}